# Genetic Algorithm for Mulicriteria Optimization of a Multi-Pickup and Delivery Problem with Time Windows

**I. Harbaoui Dridi**(1),(2) imenharbaoui@gmail.com

**R. Kammarti**(1),(2) kammarti.ryan@planet.tn

**M. Ksouri**(2) Mekki.Ksouri@insat.rnu.tn

**P. Borne**(1) p.borne@ec-lille.fr

*(1) LAGIS : Ecole Centrale de Lille, Villeneuve d'Ascq, FRANCE*

*(2) LACS : Ecole Nationale des Ingénieurs de Tunis, Tunis - Belvédère. TUNISIE*

**Abstract**: In This paper we present a genetic algorithm for mulicriteria optimization of a multi-pickup and delivery problem with time windows (m-PDPTW). The m-PDPTW is an optimization vehicles routing problem which must meet requests for transport between suppliers and customers satisfying precedence, capacity and time constraints. This paper proposes a brief literature review of the PDPTW, present an approach based on genetic algorithms and Pareto dominance method to give a set of satisfying solutions to the m-PDPTW minimizing total travel cost, total tardiness time and the vehicles number.

## 1. INTRODUCTION

Nowadays, the transport goods problem occupies an important place in the economic life of modern societies.

With the time and economic constraints implications of this problem, its resolution becomes very difficult, requiring the use of tools from different disciplines (manufacturing, information technology, combinatorial optimization, etc.)... Indeed, the process from transport systems and scheduling are becoming more complex by their large size, by the nature of their relationship dynamics, and by the multiplicity of which they are subjected.

Many studies have been directed mainly towards solving the vehicle routing problem (VRP). It's an optimization vehicle routing problem to meet travel demands. Other researchers became interested on an important variant of VRP which is the PDPTW (Pickup and Delivery Problem with Time Windows) with capacity constraints on vehicle.

The PDPTW is divided into two: 1-PDPTW (single-vehicle) and m-PDPTW (multi-vehicle).

Our object is to design a tool for m-PDPTW resolution based on genetic algorithms and Pareto dominance method to give a set of satisfying solutions to this problem minimizing total cost travelled, total tardiness time and the vehicles number.

## 2. LITERATURE REVIEW

### 2.1 VEHICLE ROUTING PROBLEM

The Vehicle Routing Problem (VRP) represents a multi-goal combinatorial optimization problem which has been the subject of much work and many variations in the literature. It belongs to the NP-hard class. . [Christofides. N and al., 1979] [Lenstra. J and al., 1981]

The Meta heuristics were also applied to solving the vehicle routing problem. Among these methods, we can include ant colony algorithms, which were used by Montamenni, R and al for the resolution of DVRP. [Montamenni. R and al., 2002]

The VRP principle is: given a depot D and a set of customers orders C = (c1, ... , Cn), to build a package routing, for a finite number of vehicles, beginning and ending at a depot. In these routing, a customer must be served only once by a single vehicle and vehicle capacity transport for a routing should not be exceeded. [Nabaa. M and al., 2007]

Savelsbergh and al have shown that the VRP is a NP-hard problem [Savelsbergh. M.P.W and al., 1995]. Since the m-PDPTW is a generalization of the VRP it's a NP-hard problem.

### 2.2 THE PDPTW: PICKUP AND DELIVERY PROBLEM WITH TIME WINDOWS

The PDPTW is a variant of VRPTW where in addition to the existence of time constraints, this problem implies a set of customers and a set of suppliers geographically located. Every routing must also satisfy the precedence constraints to ensure that a customer should not be visited before his supplier. [Psaraftis. H.N., 1983]

A dynamic approach for resolve the 1-PDP without and with time windows was developed by Psaraftis, H.N considering objective function as a minimization weighting of the total travel time and the non-customer satisfaction. [Psaraftis. H.N., 1980]

Jih, W and al have developed an approach based on the hybrid genetic algorithms to solve the 1-PDPTW, aiming to minimize combination of the total cost and total waiting time. [Jih. W and al., 1999]

Another genetic algorithm was developed by Velasco, N and al to solve the 1-PDP bi-objective in which the total travel time must be minimized while satisfy in prioritise the most urgent requests. In this literature, the method proposed to resolve this problem is based on a No dominated Sorting Algorithm (NSGA-II). [Velasco. N and al., 2006]

Kammarti, R and al deal the 1-PDPTW, minimizing the compromise between the total travel distance, total waiting time and total tardiness time, using an evolutionary algorithm with Special genetic operators, tabu search to provide a set of viable solutions. [Kammarti. R and al., 2004] [Kammarti. R and al 2005a]
This work have been extended, in proposing a new approach based on the use of lower bounds and Pareto dominance method, to minimize the compromise between the total travel distance and total tardiness time. [Kammarti. R and al 2006] [Kammarti. R and al 2007]

About the m-PDPTW, Sol, M and al have proposed a branch and price algorithm to solve the m-PDPTW, minimizing the vehicles number required to satisfy all travel demands and the total travel distance. [Sol. M and al., 1994]

Quan, L and al have presented a construction heuristic based on the integration principle with the objective function, minimizing the total cost, including the vehicles fixed costs and travel expenses that are proportional to the travel distance. [Quan. L and al., 2003]

A new metaheuristic based on a tabu algorithm, was developed by Li, H and al to solve the m-PDPTW. [Li. H and al., 2001]

Li, H and al have developed a "Squeaky wheel" method to solve the m-PDPTW with a local search. [Li. H and al., 2002]

A genetic algorithm was developed by Harbaoui Dridi, I and al dealing the m-PDPTW to minimize the total travel distance and the total transport cost. [Harbaoui Dridi. I and al., 2008]

## 3. MATHEMATICAL FORMULATION

Our problem is characterized by the following parameters:

- $N$ : Set of customers, supplier and depot vertices,
- $N'$: Set of customers and supplier vertices,
- $N^+$ : Set of supplier vertices,
- $N^-$ : Set of customers vertices,
- $K$ : Vehicle number,
- $d_{ij}$ : Euclidian distance between the vertex $i$ and the vertex $j$. If $d_{ij} = \infty$ then the road between $i$ and $j$ doesn't exist,
- $t_{ijk}$ : Time used by the vehicle $k$ to travel from the vertex $I$ to the vertex $j$,
- $[e_i, l_i]$ : Time window of the vertex $i$,
- $s_i$ : Stopping time at the vertex $i$,
- $q_i$ : Goods quantity of the vertex $i$ request. If $q_i > 0$, the vertex $i$ is a supplier; if $q_i < 0$, the vertex $i$ is a customer and if $q_i = 0$ then the vertex was served.
- $Q_k$ : Capacity of vehicle $k$,
- $i = 0..N$ : Predecessor vertex index,
- $j = 0..N$ : Successor vertex index,
- $k$: 1..K: Vehicle index,
- $Xijk = \begin{cases} 1 & \textit{If the vehicle travel from the vertex i to the vertex j} \\ 0 & \textit{Else} \end{cases}$
- $A_i$ : Arrival time of the vehicle to the vertex $i$,
- $D_i$ : Departure time of the vehicle from the vertex $i$,
- $y_{ik}$ : The goods quantity in the vehicle $k$ visiting the vertex I,
- $C_k$ : Travel cost associated with vehicle k,
- A vertex is served only once,
- There is one depot,
- The capacity constraint must be respected,
- The depot is the start and the finish vertex for the vehicle,
- The vehicle stops at every vertex for a period of time to allow the request processing,
- If the vehicle arrives at a vertex $i$ before its time windows beginning date $e_i$, it waits.

The function to minimize is given as follows:

$$Minimize\ f = \begin{cases} K, \\ \sum_{i \in N} \sum_{j \in N} max(0, D_i - l_i), \\ \sum_{i \in N} \sum_{j \in N} \sum_{k \in K} C_k d_{ijk} X_{ijk} \end{cases} \quad (1)$$

Subject to:

$$\sum_{i=1}^{N} \sum_{k=1}^{K} x_{ijk} = 1, j = 2, \ldots N \quad (2)$$

$$\sum_{j=1}^{N} \sum_{k=1}^{K} x_{ijk} = 1, i = 2, \ldots N \quad (3)$$

$$\sum_{i \in N} X_{i0k} = 1, \forall k \in K \quad (4)$$

$$\sum_{j\in N} X_{0jk} = 1, \forall k \in K \quad (5)$$

$$\sum_{i\in N} X_{iuk} - \sum_{j\in N} X_{ujk} = 0, \forall k \in K, \forall u \in N \quad (6)$$

$$X_{ijk}=1 \Rightarrow y_{jk} = y_{ik} + q_i, \forall i, j \in N; \forall k \in K \quad (7)$$

$$y_{0k}=0, \forall k \in K \quad (8)$$

$$Q_k \ge y_{ik} \ge 0, \forall i \in N; \forall k \in K \quad (9)$$

$$D_w \le D_v, \forall i \in N; \forall w \in N_i^{+}; \forall v \in N_i^{-} \quad (10)$$

$$D_0 = 0 \quad (11)$$

$$X_{ijk} = 1 \Rightarrow e_i \le A_i \le l_i, \forall i, j \in N; \forall k \in K \quad (12)$$

$$X_{ijk} = 1 \Rightarrow e_i \le A_i + s_i \le l_i, \forall i, j \in N; \forall k \in K \quad (13)$$

$$X_{ijk} = 1 \Rightarrow D_i + t_{ijk} \le (l_j - s_j), \forall i, j \in N; \forall k \in K \quad (14)$$

The constraint (2) and (3) ensure that each vertex is visited only once by a single vehicle. The constraint (4) and (5) ensure that the vehicle route beginning and finishing is the depot. The constraint (6) ensures the routing continuity by a vehicle.

(7), (8) and (9) are the capacity constraints. The precedence constraints are guaranteed by (10) and (11). The constraints (12), (13) and (14) ensure compliance time windows.

## 4. MULTI-CRITERIA EVALUATIOIN

A multi-objective problem is defined as an optimization vector problem, which seeks to optimize several components of a vector function cost.

### 4.1 PARETO DOMINANCE METHOD

A multi-criteria problem $\boldsymbol{P}$ consists of $n$ variables, $m$ inequality constraints, $p$ equality constraints and $k$ criteria whose can be formulated as follows:

$$\boldsymbol{P} \Rightarrow \begin{cases} \boldsymbol{min}\; f(x) = [f_1, f_2, f_3, \ldots\ldots f_k(x)] \\ g_i(x) \le 0 \quad \boldsymbol{i=0...m} \\ g_j(x) = 0 \quad \boldsymbol{j=0...p} \end{cases} \quad (15)$$

Therefore, it is necessary to find solutions representing a possible compromise between the criteria. The Pareto optimality concept introduced by the economist V. Pareto in the nineteenth century is frequently used [Pareto. V., 1897].

A solution is noted Pareto optimal if it is dominated by any other point in solutions space. These points are noted non-dominated solutions.

A point $X \in E$ dominates $Y \in E$ if:

$$\begin{cases} \forall i \in E, \quad f_i(x) \le f_i(y) \\ et\; \exists j, \;\; \textbf{tel que}\; f_i(x) < f_j(y) \end{cases} \quad (16)$$

Fig.1 shows an example where we seek to minimize $f_1$ and $f_2$. The points 1, 3 and 5 are not dominated. By against Point 2 is dominated by point 3, and point 4 is dominated by point 5.

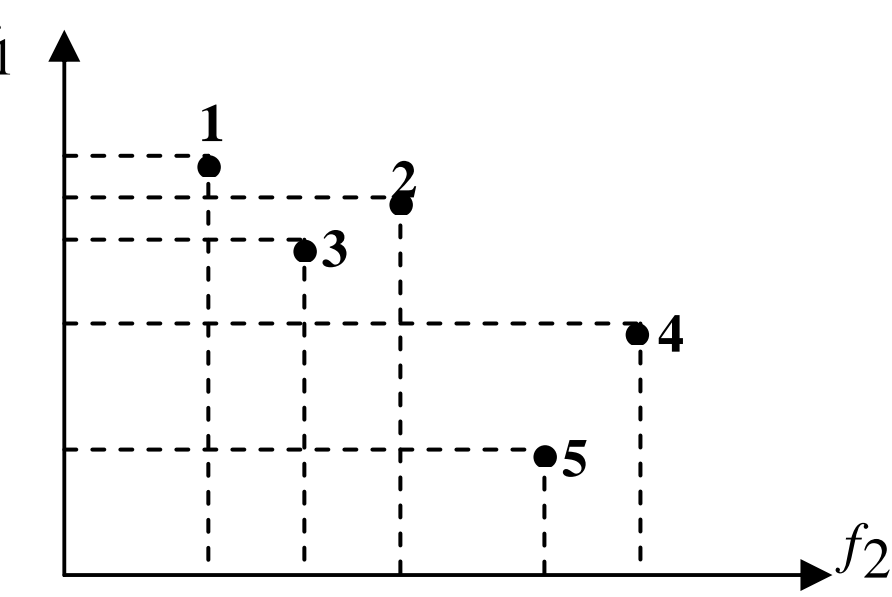

Fig.1: Dominance example

### 4.2 MATHEMATICAL MODELS

By using the parameters given above, we propose to optimize the function given by the equation (1).

Under the constraints previously noted (2)... (14).

Given that we minimize $f$ , It considers that a solution $sol_1$ dominates another $sol_2$ if $f_{11} \le f_{21}\, et\, f_{12} \le f_{22}\, et\, f_{13} \le f_{23}$. It is also considered that two solutions are not dominated by one over the other if they check a condition of following system:

$$\begin{cases} \bullet\; f_{11} > f_{21},\; f_{12} \le f_{22}\; et\; f_{31} \le f_{32} \\ \bullet\; f_{11} \le f_{21},\; f_{12} > f_{22}\; et\; f_{31} \le f_{32} \\ \bullet\; f_{11} \le f_{21},\; f_{12} \le f_{22}\; et\; f_{31} > f_{32} \\ \bullet\; f_{11} > f_{21},\; f_{12} > f_{22}\; et\; f_{31} \le f_{32} \\ \bullet\; f_{11} > f_{21},\; f_{12} \le f_{22}\; et\; f_{31} > f_{32} \\ \bullet\; f_{11} \le f_{21},\; f_{12} > f_{22}\; et\; f_{31} > f_{32} \end{cases}$$

With:

$$f_1 = K\,,$$

$$f_2 = \sum_{i\in N} \sum_{j\in N} \boldsymbol{max}(0, D_i - l_i)\,,$$

$$f_3 = \sum_{i\in N} \sum_{j\in N} \sum_{k\in K} C_k\, d_{ijk}\, X_{ijk}$$

And :

$$sol_1(f_{11}, f_{12}, f_{13})\; \textbf{et}\; sol_2(f_{21}, f_{22}, f_{23})$$

## 5. GENETIC ALGORITHM FOR MULTICRITERIA OPTIMIZATION OF M-PDPTW

### 5.1 SOLUTIONS CODING

A chromosome is a succession (permutation) vertex, which indicates the order in which a vehicle is to visit all the vertices. Fig. 2 represents the solutions under form of chromosomes.

| Vertex (i) | 0 | 5 | 8 | 2 | 6 | 4 | 3 | 10 | 7 | 9 | 1 | 0 |
|---|---|---|---|---|---|---|---|---|---|---|---|---|

Fig.2: Solution coding

The vertex "0" represents the depot.

### 5.2 GENERATION OF INITIAL POPULATION

The choice of the initial population is important because it can make a genetic algorithm more or less fast to converge towards the global optimum.

In our case, we will generate two types of populations. A first population noted $\boldsymbol{P_{node}}$, which represents all nodes to visit with all vehicles, according to the permutation list coding (Fig.2). The second population noted $\boldsymbol{P_{vehicle}}$ indicates nodes number visited by each vehicle. Knowing that $k$ varies between 1 and $\frac{\boldsymbol{N'}}{\boldsymbol{2}}$ vehicles. Fig.3 shows an individual example of $\boldsymbol{P_{vehicle}}$ with $\boldsymbol{N'}=10$.

| $V_1$ | $V_2$ | $V_3$ | $V_4$ | $V_5$ |
|---|---|---|---|---|
| 6 | 4 | 0 | 0 | 0 |

Fig.3: Individual example of $\boldsymbol{P_{vehicle}}$

Considering the population $\boldsymbol{P_{node}}$ given by Fig.2, correction procedures and $\boldsymbol{P_{vehicle}}$ population, given by Fig.3, we illustrated in Fig.4 an individual of the population $\boldsymbol{P_{node/vehicle}}$.

| $V_1$ | $C_1$ | 0 | 5 | 8 | 2 | 6 | 4 | 3 | 0 |
|---|---|---|---|---|---|---|---|---|---|
| $V_2$ | $C_2$ | 0 | 10 | 7 | 9 | 1 | 0 | | |

Fig.4: Individual of the population $\boldsymbol{P_{node/vehicle}}$

### 5.3 CROSSOVER OPERATOR

Following the generation of the initial population, we proceed to crossover phase which ensures the recombination of parental genes for train new descendants. To do this, we choose the one point crossover.

### 5.4 MUTATION OPERATOR

Mutation operator aims to choose two positions at random, within a chromosome and exchange their respective values.

### 5.5 CORRECTION PROCEDURE

The principle of correction precedence and capacity [Harbaoui Dridi, I and al., 2008] is to ensure that a customer is not visited before his supplier while respecting the vehicles capacity.

## 6. APPROACH PROPOSED TO MINIMIZE TOTA TRAVEL COST, TOTAL TARDINESS TIME AND THE VEHICLES NUMBER

After the generation of the population $\boldsymbol{P_{node/vehicle}}$, which an example is shown in Fig.3, we determine for each individual, the $f_1$, $f_2$ and $f_3$ values, which correspond to the vehicles number, the total tardiness time and the total travel cost.

We obtain the subsequent population noted $\boldsymbol{P_{pareto\text{-}dominance}}$. Fig.5 shows an individual example of this population.

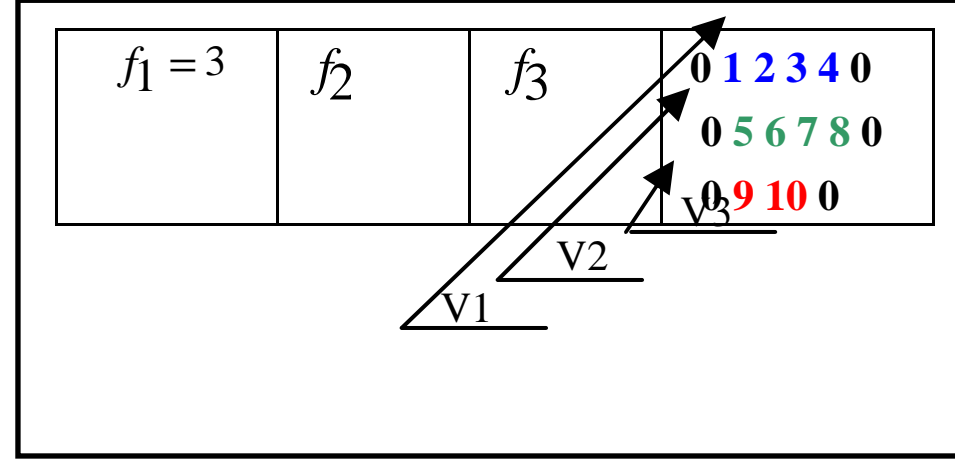

Fig.5: Individual example of $\boldsymbol{P_{pareto\text{-}dominance}}$

With:
$N'= 10$,
0: Depot

We have reproduced this result for each individual of the population $\boldsymbol{P_{node/vehicle}}$, in order to obtain thereafter the population $\boldsymbol{P_{pareto\text{-}dominance}}$.

After, we choose the population $\boldsymbol{P_{pareto\text{-}dominance}}$ according to lower values, $f_1$, $f_2$ and $f_3$, until the obtaining of all non-dominated solutions.

The following figure (Fig.6) represents the algorithm of our approach.

> ***Begin***
>
> **Step 1**: Create the initial population, (size n).
>
> **Step 2**: Fill the intermediate population $P_{node}$ (size 2n) with individuals' crossover, mutation or copy.
>
> **Step 3**: Correction procedure of Precedence and capacity.
>
> **Step 4**: Create the 2nd intermediate population $P_{node/vehicle}$ (size 2n * 2n) representing the routing of each vehicle.
>
> **While** the generation number is not reached **do**
>
> **Step 5**: Determine fitness values for each individual of the population $P_{node/vehicle}$, in order to obtain thereafter the population $P_{pareto\text{-}dominance}$
>
> **Step 6**: Sort of population $P_{pareto\text{-}dominance}$ by the minimum value of fitness (Vehicles number / Total tardiness time / Total travel cost)
>
> **Step 7**: Copy non-dominated solutions
>
> Increment the generation number
>
> **End**
> **End**

Fig.6: Approach algorithm

The procedure for determining different population is given in our work [Harbaoui Dridi, I and al., 2008].

## 7. COMPUTATIONAL RESULTS

The following table (Table.1) represents the parameters characterizing our problem.

| Id | X | Y | q | e | l | s | Succ | Pred |
|---|---|---|---|---|---|---|---|---|
| 0 | 0 | 0 | 0 | 0 | 200 | 0 | 0 | 0 |
| 1 | 85 | 56 | -20 | 41 | 67 | 11 | 0 | 5 |
| 2 | 16 | 26 | -20 | 34 | 100 | 4 | 0 | 8 |
| 3 | 57 | 26 | 20 | 69 | 124 | 2 | 10 | 0 |
| 4 | 57 | 37 | -20 | 78 | 158 | 13 | 0 | 6 |
| 5 | 71 | 69 | 20 | 62 | 64 | 12 | 1 | 0 |
| 6 | 61 | 96 | 20 | 5 | 145 | 2 | 4 | 0 |
| 7 | 22 | 17 | 20 | 27 | 81 | 1 | 9 | 0 |
| 8 | 12 | 17 | 20 | 61 | 91 | 16 | 2 | 0 |
| 9 | 92 | 85 | -20 | 95 | 142 | 18 | 0 | 7 |
| 10 | 41 | 23 | -20 | 27 | 36 | 15 | 0 | 3 |

Table.1: Processing parameters

Experimental results are given in Table.2:

| $f_1$ | $f_2$ | $f_3$ | ***Tour by vehicle*** |
|---|---|---|---|
| 2 | 0 | 30803.5 | **0 8 2 0**<br>**0 7 3 10 6 9 5 1 4 0** |
| 1 | 31.75 | 33543.9 | **0 8 2 7 3 10 6 9 5 1 4 0** |
| 1 | 0 | 51091.68 | **0 3 10 5 1 6 8 7 9 4 0** |

Table.2: Experimental results

The total tardiness time is expressed in time units.

Our approach provides a set of non-dominated solutions to ensure choice flexibility.
This set of solutions is Pareto space from which the maker will take its decision.

## 8. CONCLUSION

In this paper, we have presented our genetic approach to solve the m-PDPTW, based on Pareto dominance method. We proposed in the first part a brief literature review on the VRP, 1-PDPTW and m-PDPTW. The mathematical formulation of our problem is detailed in second part. Then, we detailed the use Pareto dominance method for determine a set of non- dominated solutions, minimizing our objective functions.